\documentclass[conference]{IEEEtran}
\IEEEoverridecommandlockouts
\usepackage{cite}
\usepackage{amsmath,amssymb,amsfonts}
\usepackage[pagebackref=true,breaklinks=true,bookmarks=false]{hyperref}
\usepackage{algorithmic}
\usepackage{graphicx}
\usepackage{textcomp}
\usepackage{xcolor}
\usepackage{hyperref}
\usepackage{epsfig}
\usepackage{graphicx}
\usepackage{amsmath}
\usepackage{amssymb}
\usepackage{booktabs}
\usepackage{adjustbox}
\usepackage{enumitem}
\usepackage{multirow}
\usepackage{marvosym}
\usepackage{caption}

\begin{document}

\title{Predicting Parkinson's Disease with Multimodal Irregularly Collected Longitudinal Smartphone Data\\}

\author{\IEEEauthorblockN{Weijian Li\IEEEauthorrefmark{1},
Wei Zhu\IEEEauthorrefmark{1},
E. Ray Dorsey\IEEEauthorrefmark{2,3}, and
Jiebo Luo\IEEEauthorrefmark{1}}
\IEEEauthorblockA{\IEEEauthorrefmark{1}Department of Computer Science, 
University of Rochester, Rochester, NY, USA}
\IEEEauthorblockA{\IEEEauthorrefmark{2}Center for Health + Technology and Department of Neurology, University of Rochester, Rochester, NY, USA}
\IEEEauthorblockA{Email:\IEEEauthorrefmark{1}\{wli69, wzhu15, jluo@cs.rochester.edu\}, \IEEEauthorrefmark{2}\{ray.dorsey@chet.rochester.edu\}}}

\maketitle

\begin{abstract} Parkinson’s Disease is a neurological disorder and prevalent in elderly people. Traditional ways to diagnose the disease rely on in-person subjective clinical evaluations on the quality of a set of activity tests. The high-resolution longitudinal activity data collected by smartphone applications nowadays make it possible to conduct remote and convenient health assessment. However, out-of-lab tests often suffer from poor quality controls as well as irregularly collected observations, leading to noisy test results. To address these issues, we propose a novel time-series based approach to predicting Parkinson's Disease with raw  activity test data collected by smartphones in the wild. The proposed method first synchronizes discrete activity tests into multimodal features at unified time points. Next, it distills and enriches local and global representations from noisy data across modalities and temporal observations by two attention modules. With the proposed  mechanisms, our model is capable of handling noisy observations and at the same time extracting refined temporal features for improved prediction performance. Quantitative and qualitative results on a large public dataset demonstrate the effectiveness of the proposed approach.

\end{abstract}

\begin{IEEEkeywords} Parkinson's Disease, Multimodal data, Smartphone, Neural Ordinary Differential Equations
\end{IEEEkeywords} 

\section{Introduction} As the second most prevalent chronic neurodegenerative movement disorder in the world, Parkinson's Disease (PD) is on a remarkable increase over the past years~\cite{rossi2018projection} and continuously posing severe threats to patient health with symptoms such as stress, tremor, degradation of memory and physical activities. Medical treatments are available to mitigate the PD symptom effects. Thus, timely and correct diagnosis of PD is crucial for early interventions prior to serious deterioration. A typical way of diagnosing the Parkinson's Disease is through in-person assessment with clinicians. However, PD symptoms could be variable over time~\cite{becker2002early} which influences the onsite diagnosis quality given sparsely obtained health condition records and potential unobservable health condition changes. In addition, diagnosis by clinicians are usually subjective and difficult to calibrate.

\begin{figure}[t]
	\centering
	\resizebox{0.7\columnwidth}{!}{\includegraphics{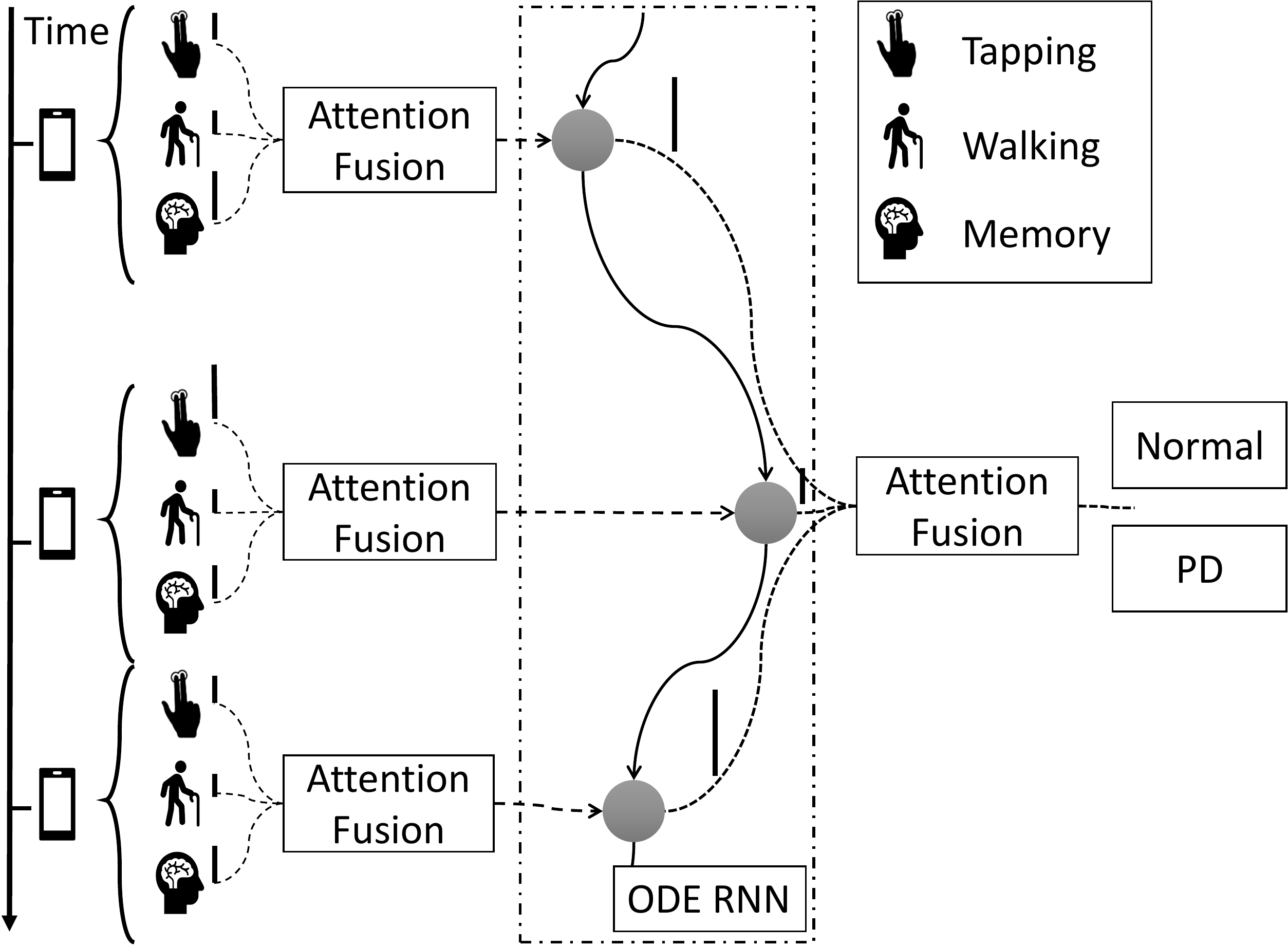}}
	\caption{Illustration on predicting the Parkinson's Disease (PD) as well as a brief overview of the proposed approach. A time-series based learning approach with attention mechanisms on both temporal and modality features, adaptively aggregates multimodal activity information for final PD prediction.}
	\label{fig:demo}
	\vspace{-5mm}
\end{figure}
Recent studies~\cite{bot2016mpower,zhang2019pdvocal,zhan2016high} develop device-based software that include remote health measurements. The \textit{mPower} study~\cite{bot2016mpower}, for example, proposes a smartphone-based App that provides clinical related PD tests, which involve interactions with the participants and can be conducted outside clinics and at any time. 
Such remote health access approaches show an opportunity for timely PD diagnosis as well as improving disease understanding with the enriched longitudinal quantitative records~\cite{schwab2019phonemd}. The nature of the obtained device signals could also facilitate normalized objective measurements. However, the out-of-lab measurements pose new challenges: (1) the uncontrollable test time-points and the combination of test subjects lead to irregularly distributed results in the temporal dimension; (2) the self-reported test results lack quality control which may introduce noisy observations and affect the overall prediction performance. These are common difficulties when dealing with real-world multidimensional time-series data~\cite{schneider2001analysis,cui2018deep}, especially in the medical domain~\cite{clifford2016classification,johnson2016mimic}.

\begin{figure*}[htbp]
	\centering
	\resizebox{1.8\columnwidth}{!}{\includegraphics{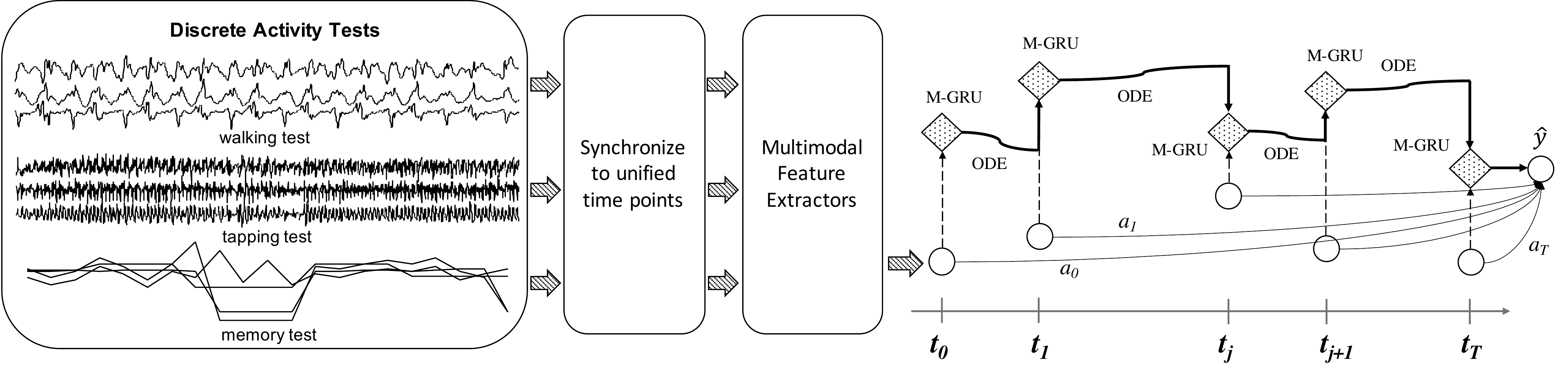}}
	\caption{Overview of the proposed framework. Our model mainly consists of three modules:~(1) {\it multimodal feature extractor}, (2) {\it temporal encoder with multimodal attention} and (3) {\it embedding Self-Attention Pooling}.}
	\label{fig:framework}
	\vspace{-2mm}
\end{figure*}
Several methods have been proposed to tackle the irregularly longitudinally distributed samples~\cite{prince2018multi,zhang2019cpm,che2018recurrent,tandata,baytas2017patient,chen2018neural,rubanova2019latent,de2019gru}. Among them, the Neural Ordinary Differential Equations (ODEs)~\cite{chen2018neural,rubanova2019latent,de2019gru,shi2020cubic} are a group of continuous-time models with a series of hidden states in the latent space. The observed fix-interval time-series signal with possibly missing values can then be modeled by continuous latent space representations. Given the function dynamics and a numerical ODE solver, each hidden state can be computed, representing the latent trajectory. However, in the medical field, present Neural ODE based methods focus on in-hospital collected data, e.g. patient ICU measures~\cite{clifford2016classification} and EHR records~\cite{johnson2016mimic}, the effective way to deal with noisy self-reported multimodal data in the wild remains unclear. 

To address the above issues, we present a novel end-to-end deep-learning based model for predicting Parkinson's Disease with self-reported multimodal smartphone data collected in the wild. To be specific, the proposed method first extracts feature representations from different modalities using mode-related encoders. An ODE based time-series encoder is then introduced to map the observed signals into a latent space for continuous modeling. Finally, a state-wise self-attention mechanism is proposed to learn aggregate local features for the prediction task and, more importantly, for better model interpretability important to clinical practice. 

In summary, our main contributions are three folds:
\begin{itemize}
	\item We predict Parkinson's Disease based on sporadically-observed activity data in the wild collected from smartphones with Neural Ordinary Differential Equations (ODEs). To our best knowledge, this is the first attempt at time-series prediction of Parkinson's Disease in an uncontrolled environment. The proposed model has the potential to be adapted to similar tasks.
	
	\item We synchronize discrete observations into unified time-points and extract valuable multimodal representations from noisy data with a multimodal attention mechanism.

	\item We aggregate temporal observations with a self-attention mechanism for an enriched joint local and global representation as well as improving the interpretability for clinical practice.

\end{itemize}

\section{Methods} 
\subsection{Overview} 
\noindent \textbf{Problem Definition} Our work is based on the data collected by a large-scale Parkinson's Disease study named mPower~\cite{bot2016mpower}, which contains activity tests conducted by the participants through smartphones. Besides, participants also report their PD diagnosis status. Thus, given this remotely collected dataset, our goal is to correctly classify each participant as a PD patient or a non-PD patient. The overview of the proposed framework can be found in Figure~\ref{fig:framework}.

\subsection{Multimodal Feature Extractor} The goal of the multimodal feature extractor is to encode raw activity test results into enriched feature embeddings as the model inputs for end-to-end learning. Considering the sequential data samples conducted in each of the activity tests, e.g. the accelerometer sequence generated when the participant is walking, we decide to adopt the widely adopted Temporal Convolutional Networks (TCN)~\cite{bai2018empirical} as the raw feature extractors. In detail, given an observed test sequence $x_t^m=\{x_0,...,x_{L_{mt}}\}$ with length $L_{mt}$ from modal $m \in M$ at observation time $t \in T$ 
, each layer of TCN applies a dilated convolution on the test sequence with a focus on the longitudinal causal relationship:
\begin{equation}
	F(s^l_n)= \sum_{i=1}^{k-1} f(i) s^{l-1}_{n-d* i}
	\label{eqn:tcn_conv}
\end{equation}
where $d$ is the dilation rate, and $k$ is the filter size. Then the $n$th element in the $l$th layer is computed with the convolutional kernel $f$ applied on the $n-d*i$ elements in the $l-1$th layer. In this way, we obtain the embeddings $v_t =\{v_t^0,...v_t^m\}$ with $v_t^m\in\mathbb{R}^{D \times 1}$ for each modality $m$ at observation time $t$.

\subsection{Temporal Encoder with Multimodal Attention} Traditional ways of handling irregularly sampled data rely on direct aggregations or data imputation strategies. However, these methods are likely to cause information loss at important sequential and continuous signal changes, which are crucial for health condition analysis. Given the extracted and aligned multimodal features at each time point, we propose to model continuous PD symptom changes in the latent space through the Neural Ordinary Differential Equations (ODEs)~\cite{chen2018neural,rubanova2019latent}. 

Recall that for ODEs, time-series is represented by a latent trajectory determined by the initial state $h_0$. Given the observed time points $t_0,t_1,...,t_T$ and an initialized state $h_0$, an ODE solver computes $h_1,...,h_T$ representing the hidden states for each time point. Formally, 
    \begin{gather}	\label{eqn:ode}
        h_0 \sim p(h_0) \\
    	h_1,...,h_T = \mbox{ODESolve}(h_0, f, \theta_f, t_0,...,t_T)
	\end{gather}
where function $f$ produces the gradient $\frac{\partial h(t)}{\partial t}=f(h(t),\theta_f)$ which is parameterized with a neural network. Each hidden state $h_t$ is then obtained by integrating the gradient through time, which is achieved by an ODESolver. To incorporate the observations at each time point $t$ and adjust the latent trajectory accordingly, hidden states are updated by a network, e.g. an RNNCell:
\begin{equation}
        h_t = \mbox{RNNCell}(h_t', u_t)
	\label{eqn:odernn}
\end{equation}
where $h_t'$ is the hidden state before the update and $u_t$ is the observation features at current time $t$. 

A simple way to construct input $u_t$ is by directly concatenating the embeddings $v_t$ extracted from the original observations. However, since participants are free to choose the types of activity tests to perform at home, the observed test results at each time-point are usually incomplete, resulting in some of the modal features being missing. Therefore, we attach each of the modality features with a binary mask $v_t'^{m}$ with the same feature length, indicating its observation status: $v_t^m \leftarrow [v_t^m \cdot v_t'^{m}]$.

In addition, different modalities may contribute differently to the final PD prediction due to abnormal or noisy test results. Given the hidden state $h_t'$, a valid modality test should share common representations that consider the similar semantics, e.g. the identity of the same participant. Based on this observation, we propose to integrate an attention mechanism inside the GRU cell, named M-GRU, to further adaptively learn an aggregation function by assigning a weight to each of the modalities: 
    \begin{gather}	\label{eqn:att_modal}
        u_t' = \sum_{m=1}^{M} a_m * v_t^m
	\end{gather}
where
    \begin{gather}	\label{eqn:att_modal_2}
        a_m = \frac{\mbox{exp}\{e_m\}}{\sum_{m=1}^{M} \mbox{exp}\{e_m\}}\\
        e_m = w_m^T\mbox{tanh}(W_{hm}* h_t' + W_{vm}* v_t + b_m)
	\end{gather}
where $w_m$, $W_{hm}$, $W_{vm}$ and $b_m$ are learnable parameters for computing the transformed representaion $e_m$. In this paper, we adopt the GRU unit as the RNN cell for updating the hidden state $h_t$. The state-wise input is a concatenation of the original inputs and the attended representations: $u_t = [v_t \cdot u_t']$. Therefore the updating function can be written as follows:
    \begin{gather}	\label{eqn:gru}
        z_t = \sigma(W_z*[h_t',v_t])\\
        r_t = \sigma(W_r*[h_t',v_t])\\
        \tilde{h_t} = \sigma(W_g*[r_t*h_t',u_t])\\
        h_t = (1-z_t)*h_t'+z_t*\tilde{h_t}
	\end{gather}
where $W_z$, $W_r$ and $W_g$ are learnable parameters, $\sigma$ is the hyperbolic tangent function.

\begin{figure}[!t]
	\centering
	\resizebox{0.8\columnwidth}{!}{\includegraphics{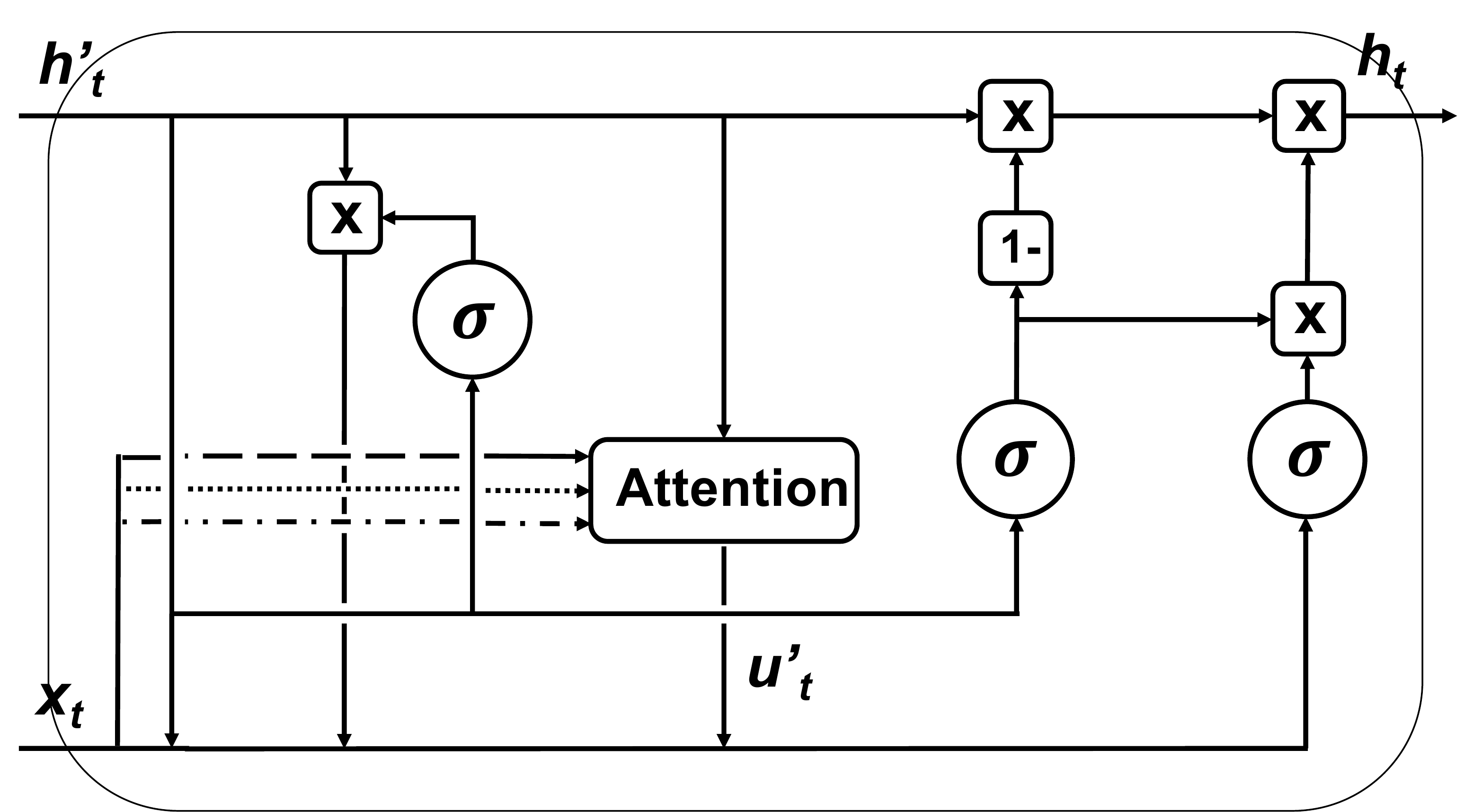}}
	\vspace{0mm}
	\caption{Illustration on the multimodal attention mechanism based on a GRU cell. At each observation point, each of the modality is attended by the incoming hidden state $h_t'$ prior to update. The attended value $u_t'$ is then concatenated with hidden state $h_t'$ and observational inputs $x_t$ for the following $h_t$ computation. $\sigma$ represents the hyperbolic tangent function.}
	\label{fig:multimodal_attention}
	\vspace{-2mm}
\end{figure}

\subsection{Embedding Self-Attention Pooling} Different from the observations in controlled environments, e.g. ICUs in hospitals, self-reported test results suffer from poor quality control. Adopting one single state as the user representation for prediction can be easily biased by certain noisy observations. To increase our model's robustness and extract raw symptom clues from each modality, we adopt a self-attention mechanism~\cite{ilse2018attention} on all the encoded modality features at each step $v_t$ to form a time-wise global representation:
    \begin{gather}	\label{eqn:att_time}
        h = \sum_{t=1}^{T} a_t * v_t
	\end{gather}
where
    \begin{gather}	\label{eqn:att_time_2}
        a_t = \frac{\mbox{exp}\{w^T\mbox{tanh}(Wv_k^T)\}}{\sum_{t=1}^{T} \mbox{exp}\{w^T\mbox{tanh}(Wv_k^T)\}}
	\end{gather}
where $v_t$ is a concatenation of the extracted modal features $v_t^m$, while $w$ and $W$ are learnable parameters. We then concatenate the time-wise representation with the last hidden state from the temporal encoder as user representation logits. The final prediction is obtained by applying the sigmoid function on the transformed logits:

\begin{equation} \label{eqn:pred}
	\hat{y}=\mbox{Sigmoid}(w^T[h\cdot h_T]+b)
\end{equation}

\subsection{Training} We adopt the standard binary cross-entropy loss on the predicted logit $\hat{y}$ and the target label $y$:
\begin{equation}
	\mathcal{L}= -\frac{1}{N}\sum_{i=1}^{N} y_ilog(\hat{y_i})+(1-y_i)log(1-\hat{y_i})
\end{equation}

\section{Experiment}
\subsection{Dataset Description} We first give a brief review on the mPower dataset~\cite{bot2016mpower}\footnote{\url{https://www.synapse.org/\#!Synapse:syn4993293/wiki/247859}}. It contains four types of PD-related activity test results from the participants conducted on their smartphones. In this study, we adopt three of them and leave the more complicated voice signals for future work:

\noindent \textbf{Tapping Test} It measures the impaired ﬁnger dexterity and tapping speed which are common signs of Parkinson's Disease. In this test, participants are asked to place their smartphone on a flat surface and use the two fingers from the same hand to tap two buttons shown on their screen alternatively for 20 seconds.

\noindent \textbf{Walking Test} It evaluates participant's gait and balance. During this test, participants need to carry the smartphone in the pocket and walk out-bounds, stand still then walk back.

\noindent \textbf{Memory Test} It focuses on evaluating participant's short-term spatial memory. During this test, participants are shown an illuminated pattern on their smartphone screen and asked to replicate the pattern by touching the corresponding places in the correct order.

Each of the tests also asks participants to choose their medication points when conducting the test, namely: \textit{Immediately before Parkinson medication}, \textit{Just after Parkinson medication (at your best)}, \textit{I don't take Parkinson medications}, and \textit{Another time}. For the tapping and walking test, we adopt the accelerometer readings which contain $(x, y, z)$ coordinate sequences in Gs. For the memory test, we use both the tapping response sequences, the corresponding targets and the time spent for the response.

\begin{table}[!t]
  \centering
    \caption{Statistics of the dataset after preprocessing and synchronizing the three modalities into common time periods. $\pm$ represents a standard deviation following the mean value.}
  \resizebox{0.9\linewidth}{!}{
      \begin{tabular}{l|cc}
        \toprule
                \textbf{Properties} & \multicolumn{2}{c}{\textbf{Values}}  \\
        \midrule
                Samples (\#)  & \multicolumn{2}{c}{1,236} \\
                Gender: Male \& Female (\%) & \multicolumn{2}{c}{67.7\% \& 32.3\%}  \\
                PD \& Non-PD (\%)  & \multicolumn{2}{c}{62.4\% \& 37.6\%} \\
                Age  (\#) & \multicolumn{2}{c}{60.61 $\pm$ 8.76}  \\
        \midrule
                Total tests (\#) \& Missing rate (\%) & \multicolumn{2}{c}{122,790 \& 57.6\%}  \\
                Sequence length per ID (\#) & \multicolumn{2}{c}{13.75 $\pm$ 23.51} \\
                Walking test per ID (\#) & \multicolumn{2}{c}{6.49 $\pm$ 14.21} \\
                Tapping test per ID (\#) & \multicolumn{2}{c}{13.26 $\pm$ 20.62} \\
                Memory test per ID (\#) & \multicolumn{2}{c}{1.73 $\pm$ 6.33} \\
                
        \bottomrule
      \end{tabular}
  }
\label{tab:stat}
\vspace{-1mm}
\end{table}

\subsection{Data Preprocessing}
The mPower dataset is collected by the participants outside hospitals with limited quality control. To achieve our goal of multimodal time-series analysis, careful data pre-processing is crucial to remove noisy signals. Due to a large variation on the time when the test is conducted, temporally synchronize test results across different modalities to obtain multimodal observations at unified time points is also needed. We preprocess the collected data as follows:

\noindent\textbf{Accelerometer Sequences} For the tapping and walking tests, we adopt the accelerometer readings from the smartphone which contain sequences of $(x, y, z)$ coordinates. Each of the sequences is first processed with the low-pass filters~\cite{badawy2018automated} to remove the gravitational component. Since the tests are conducted in a highly uncontrolled environment, noisy observations, e.g. no tapping or not standing still, need to be removed. A publicly available\footnote{ \url{https://github.com/deepcharles/ruptures}} change point detection algorithm~\cite{truong2020selective} is then applied on the processed signal to segment the potential movements of interest. The longest segmented sequence with a signal standard deviation above a predefined threshold is extracted as the final observed sequence.

\noindent\textbf{Memory Records} For the memory tests, we adopt each participant's actual button-tapping sequences and the corresponding target button sequences through time. If the participant play the memory game multiple times during the test, we concatenate the tests sorted by time. Game scores generated by the App are also attached to each of the touches in the game with a four dimensional representation for each touch: $(time, actual, target, score)$.

\noindent\textbf{Time Synchronization} Notice that a different PD medication point may influence the test performance, e.g. \textit{just before medication} is worse than \textit{at your best}. To remove this effect, we first group the records by participant IDs and the medication point when the test is conducted. The test records with the \textit{Another time} medication status is not used due to its ambiguous representation. The unique combinations of \textit{Participant ID + Medication Point} are considered as the new unique IDs. To construct multimodal representations for each ID at unified time periods, the obtained records from different modalities within 24 hours are then grouped together. If there are duplicate records in the same time period, we sort them by the average observation time of the three modalities, and only the last observed one is kept.

\noindent\textbf{Other Preprocessings} Similar to previous studies~\cite{schwab2019phonemd,prince2018multi}, we remove participants with ages below 45 who are less likely with PD symptoms. Participants perform less than 5 tests in total are also not included in the study. In the end, we obtain 1,236 samples containing their synchronized multimodal sequences. Detailed dataset statistics can be found in Table~\ref{tab:stat}.

\subsection{Methods for Comparison} We compare the proposed method with 
six
baseline models. Three of them are traditional methods while the other 
three
are deep learning based including the state-of-the-art time-series analysis
methods RNN+$\Delta t$, GRU-D, and ODE-RNN.

\begin{table}[!t]
  \centering
    \caption{Evaluation results with 5-fold cross-validation. $\pm$ represents a standard deviation following the mean value of the five folds.}
    \resizebox{0.9\linewidth}{!}{
      \begin{tabular}{lccc}
        \toprule 
        \cmidrule{1-4}
                Method & AUC & AUPR & F1 \\
        \cmidrule{1-4}
                LR~\cite{kleinbaum2002logistic} & 0.556 $\pm$ 0.028 & 0.665 $\pm$ 0.058 & 0.521 $\pm$ 0.071 \\
                SVM~\cite{suykens1999least}  & 0.547 $\pm$ 0.057 & 0.657 $\pm$ 0.049 & 0.697 $\pm$ 0.022 \\
                XGBoost~\cite{chen2016xgboost} & 0.631 $\pm$ 0.042 & 0.726 $\pm$ 0.030 & 0.730 $\pm$ 0.029 \\
        \midrule
                RNN~\cite{mikolov2010recurrent}+$\Delta$t & 0.726 $\pm$ 0.026 & 0.811 $\pm$ 0.035 & 0.771 $\pm$ 0.012 \\
                GRU-D~\cite{che2018recurrent} & 0.754 $\pm$ 0.030 & 0.827 $\pm$ 0.033 & 0.788 $\pm$ 0.023 \\
                ODE-RNN~\cite{rubanova2019latent} & 0.767 $\pm$ 0.022 & 0.845 $\pm$ 0.031 & 0.797 $\pm$ 0.023 \\
                \textbf{Proposed} & \textbf{0.793 $\pm$ 0.024} & \textbf{0.865 $\pm$ 0.028} & \textbf{0.816 $\pm$ 0.021} \\
        \bottomrule
      \end{tabular}
      \label{tab:eval_results}
      }
\vspace{-3mm}
\end{table}

\begin{itemize}
	\item \textbf{LR}~\cite{kleinbaum2002logistic}: We leverage a standard logistic regression classifier for binary classification.
	\item \textbf{SVM}~\cite{suykens1999least}: We adopt a standard Support Vector Machine classifier with the RBF kernel for comparison.
	\item \textbf{XGBoost}~\cite{chen2016xgboost}: It stands for Extreme Gradient Boosting which is a tree-based boosting algorithm.
	\item \textbf{RNN+$\Delta$t}: We concatenates intervals $\Delta$t to the input features and feed them into a standard RNN model~\cite{mikolov2010recurrent}. The last hidden state is used for the final prediction.
	\item \textbf{GRU-D}~\cite{che2018recurrent}: The GRU-D model is also designed for modeling trajectory changes with a hidden state exponentially decay through time. In addition, it concatenates observational masks and time intervals between the observations as additional clue into the inputs.
	\item \textbf{ODE-RNN}~\cite{rubanova2019latent}: The ODE-RNN model focuses on the continuous latent space trajectory modeling which captures inter-observation changes with ODE and at-observation hidden state updates with a GRUCell.
\end{itemize}

\subsection{Experimental Settings} Baseline methods LR and SVM are adopted from the Scikit-Learn library. The XGBoost algorithm is adopted from its publicly available python version. Since LR, SVM and XGBoost algorithms are not designed for handling irregular time-series inputs, we take the average of the features from all the steps as the global representation and feed them into these classifiers. For each step, we concatenate three modalities as a combined representation. The remaining methods use the same inputs and the same settings for the feature extractor and final prediction network. During training, the Adam optimizer is used with a learning rate initialized as 0.01 and decay by 0.96 for each epoch. All experiments are conducted with 5-fold cross-validation. Area Under the Curve (AUC), Area Under the Precision-Recall Curve (AUPR) and F1 scores are used as the evaluation metrics. We report the average and standard deviation values obtained from cross-validation. The proposed model is implemented in PyTorch and experimented on a single NVIDIA Geforce GTX 1080Ti GPU.

\begin{figure}[t]
	\centering
	\resizebox{0.8\columnwidth}{!}{\includegraphics{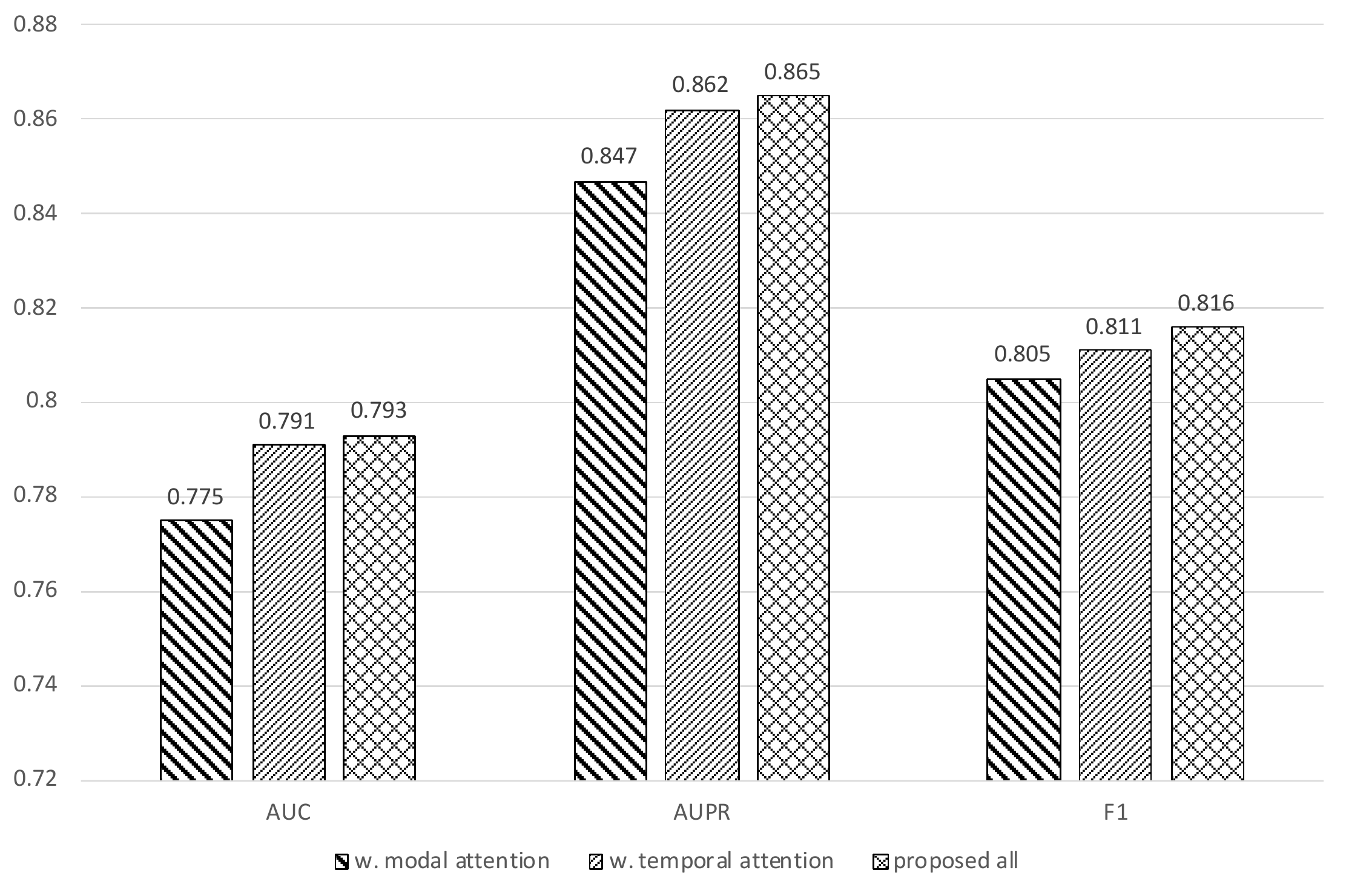}}
	\caption{Comparison on the performance of the proposed attention mechanisms.}
	\label{fig:chart_att}
	\vspace{-5mm}
\end{figure}

\begin{figure}[htbp]
	\centering
	\resizebox{0.9\columnwidth}{!}{\includegraphics{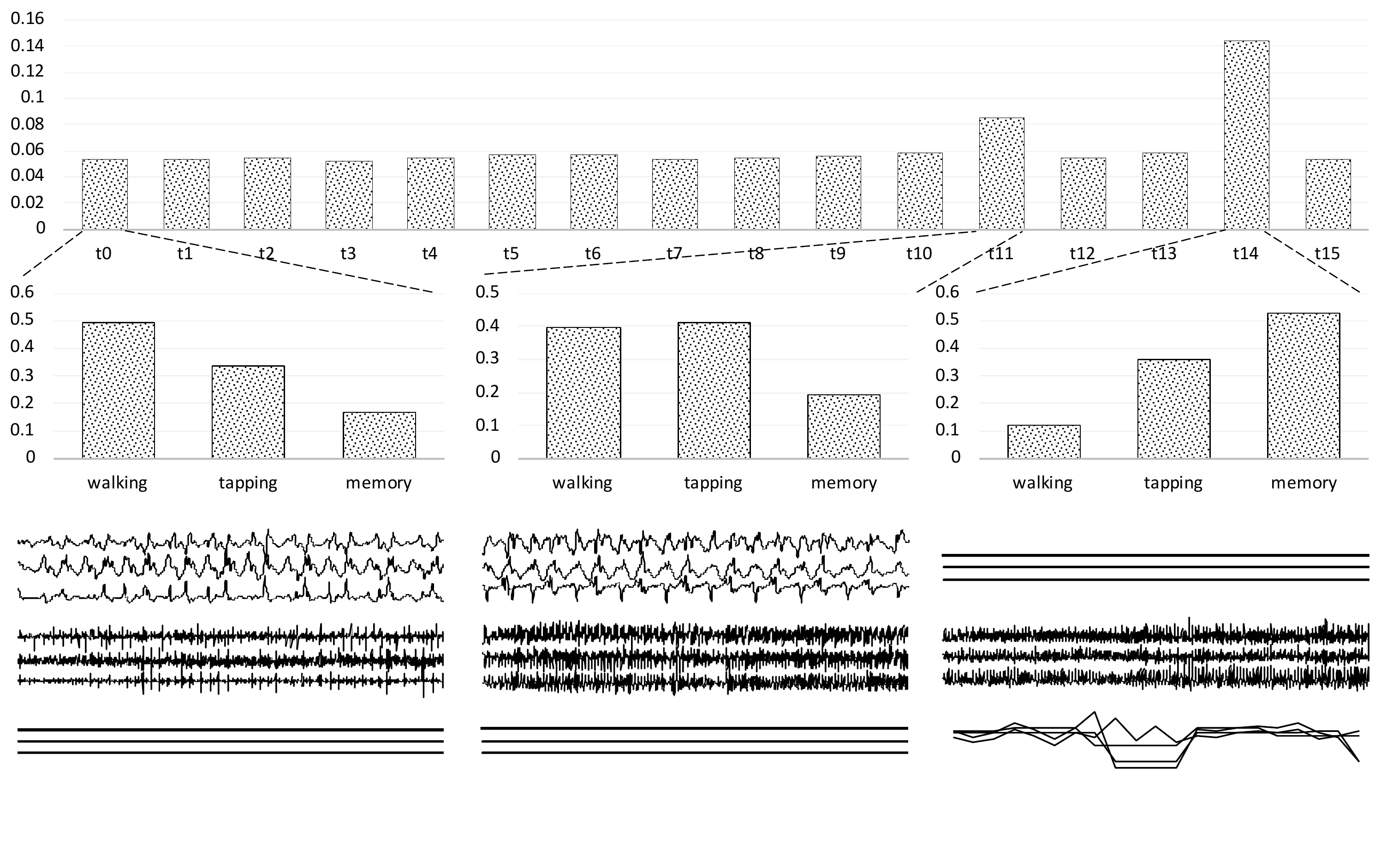}}
	\vspace{-5mm}
	\caption{Visualization of the obtained attention values for a sequence of 16 ($t_0-t_{15}$) tests from a participant. Top chart: the attention weights on each of the temporal observations. Three middle charts: exemplar attention weights for three modalities at time $t_0$, $t_{11}$ and $t_{14}$. Bottom charts: raw signals for walking, tapping and memory tests at the three observation times. Horizontal lines represent no test result for the modality.}
	\label{fig:attention_visual}
	\vspace{-3mm}
\end{figure}

\subsection{Quantitative Evaluations} 
\noindent\textbf{Overall performance} As shown in Table~\ref{tab:eval_results},  the proposed method in general achieves the best performance compared to all the strong baseline methods, e.g. RNN$+\Delta t$ and ODE-RNN, over which our method achieves 10\% and 5\% performance gain in AUC respectively, demonstrating its effectiveness in the PD prediction task. LR, SVM and XGBoost perform worse than the deep-learning based methods. We think there are two main reasons. First, the features used for these methods are losing important temporal information by aggregating through time with average pooing. A trajectory that records a participant's test performance is shown to be helpful for extracting the long-term PD symptom patterns that benefit our prediction task. The high model capacity of the deep learning based methods also provides higher power for dealing with high dimensional features. Encoding raw test signals and includes them into end-to-end learning also helps learn rich feature inputs. For the deep learning based methods, RNN$+\Delta t$ performs better than non-deep-learning methods, indicating the benefits of time-series pattern learning as well as the integration of time-interval information. In addition to the updates at each observation, GRU-D also considers the dynamic changes between the observations by introducing an exponential hidden state decay mechanism that constructs a temporal relationship with respect to the time intervals. ODE-RNN further expands this idea by leveraging the ODE solvers to compute the derivatives for hidden state changes. Different from a predefined decay, ODE models are more flexible for handling continuous state changes with arbitrary time-intervals, leading to enriched latent space trajectory representation. The proposed method achieves the best results across all three evaluation metrics. 

\noindent\textbf{On proposed attention modules} From Figure~\ref{fig:chart_att}, we can see all  the models with the proposed attention mechanism separately achieve better performance than the previous methods. Combining the two proposed attention mechanisms together, we achieve the best results which indicate a mutual improvement effect. When looking more closely, we find that our model with temporal attention brings the most improvements  with a similar result when adopting both of the attention mechanisms. Recall that the temporal attention aggregates embedded multimodal features at each observation time to a unified representation. In this way, we believe our model not only preserves the original local modal representations but also learns to extract the most informative ones that provides our model with extra knowledge for decision making.



\subsection{Qualitative Evaluation} 
\noindent\textbf{Attention visualization} To further examine the effect of the proposed method, we visualize the attention weights learned by each of the attention modules. As shown in Figure~\ref{fig:attention_visual}, from a global temporal point of view, each step is assigned an attention value, with the largest one at time $t_14$ and the second at time $t_11$. This difference indicates our model is looking for certain patterns from each of the observations. Looking more into the details, we highlight three of the representative observation time points, namely $t_{14}$, $t_{11}$ and a lower attended $t_0$. For $t_{14}$, we find that the memory test is being paid the most attention, following by the tapping test. The lowest weight is given to the walking test where no test result is presented. We consider some reasons are behind the memory test's highest weight. One is that the memory test is less affected by potential noise because the task itself takes much fewer body movements than walking and tapping (continuously tapping the screen). Another is that the performance of the memory test is easier to quantify by directly comparing participant's responses (actual tapping sequence, response speed) to clear targets (target tapping sequence, faster response speed),  which helps measure the health status and improve the PD prediction accuracy. For $t_{11}$ and $t_0$, we find in both cases the memory test is not conducted. Instead, our model focuses differently on tapping and walking. Yet notice that our model could be biased to memory tests if the existence of memory test is actually a reflection of PD existence. Future work could be directed on analyzing data bias problems for better generalizability. Looking into the signals, we find a more intense tapping sequence in $t_{11}$ than the one in $t_0$ which may contain richer behavior patterns for analysis. 


\section{Conclusion} In this paper, we present a novel time-series based deep learning approach to Parkinson's Disease prediction based on remotely and irregularly collected data from smartphones. Different from previous methods, we synchronize discrete observations to unified observational time points to construct multimodal time-series representations using the Neural Ordinary Differential Equations. Two proposed attention mechanisms adaptively learn important features from noisy signals in both the temporal and modality dimensions. Insights and improved quantitative and qualitative results on a large public dataset demonstrate the effectiveness of the proposed approach.

\section{Acknowledgement} Research reported in this publication was supported by the National Institute Of Neurological Disorders And Stroke of the National Institutes of Health under Award Number P50NS108676. The content is solely the responsibility of the authors and does not necessarily represent the official views of the National Institutes of Health. We also thank for the support from NSF through award IIS-1722847.


{\small
\bibliographystyle{IEEEtran}
\bibliography{main}
}

\end{document}